\documentclass{article}

\usepackage{arxiv}

\usepackage[utf8]{inputenc} 
\usepackage[T1]{fontenc}    
\usepackage{hyperref}       
\usepackage{url}            
\usepackage{booktabs}       
\usepackage{amsfonts}       
\usepackage{nicefrac}       
\usepackage{microtype}      
\usepackage{lipsum}

\usepackage{graphicx}
\usepackage{algorithm}
\usepackage{algorithmic}
\usepackage{booktabs}
\usepackage{amsfonts}
\usepackage{xcolor}
\usepackage{multirow}
\usepackage{verbatim}

\title{A Neural Approach to Irony Generation}

\author{
  Mengdi Zhu, Zhiwei Yu, Xiaojun Wan\\
  Peking University\\
\{1600012990,yuzw,wanxiaojun\}@pku.edu.cn
}

\begin{document}
\maketitle

\begin{abstract}
Ironies can not only express stronger emotions but also show a sense of humor. With the development of social media, ironies are widely used in public. Although many prior research studies have been conducted in irony detection, few studies focus on irony generation. The main challenges for irony generation are the lack of large-scale irony dataset and difficulties in modeling the ironic pattern. In this work, we first systematically define irony generation based on style transfer task. To address the lack of data, we make use of twitter and build a large-scale dataset. We also design a combination of rewards for reinforcement learning to control the generation of ironic sentences. Experimental results demonstrate the effectiveness of our model in terms of irony accuracy, sentiment preservation, and content preservation\footnote{Our data and code are released at https://github.com/zmd971202/IronyGeneration.}.
\end{abstract}

\section{Introduction}
The irony is a kind of figurative language, which is widely used on social media \cite{DBLP:conf/semeval/GhoshLVRSBR15}. The irony is defined as a clash between the intended meaning of a sentence and its literal meaning \cite{DBLP:conf/eacl/MoriceauKBPBA17}. As an important aspect of language, irony plays an essential role in sentiment analysis \cite{DBLP:conf/semeval/RosenthalRNS14,DBLP:conf/semeval/GhoshLVRSBR15} and opinion mining \cite{DBLP:journals/ftir/PangL07,sarmento2009automatic}. 

Although some previous studies focus on irony detection, little attention is paid to irony generation. As ironies can strengthen sentiments and express stronger emotions, we mainly focus on generating ironic sentences. Given a non-ironic sentence, we implement a neural network to transfer it to an ironic sentence and constrain the sentiment polarity of the two sentences to be the same. For example, the input is ``I hate it when my plans get ruined" which is negative in sentiment polarity and the output should be ironic and negative in sentiment as well, such as ``I like it when my plans get ruined". The speaker uses ``like" to be ironic and express his or her negative sentiment. At the same time, our model can preserve contents which are irrelevant to sentiment polarity and irony. According to the categories mentioned in \cite{DBLP:conf/semeval/HeeLH18}, irony can be classified into 3 classes: verbal irony by means of a polarity contrast, the sentences containing expression whose polarity is inverted between the intended and the literal evaluation; other types of verbal irony, the sentences that show no polarity contrast between the literal and intended meaning but are still ironic; and situational irony, the sentences that describe situations that fail to meet some expectations. 
As ironies in the latter two categories are obscure and hard to understand, we decide to only focus on ironies in the first category in this work. For example, our work can be specifically described as: given a sentence ``I hate to be ignored", we train our model to generate an ironic sentence such as ``I love to be ignored". Although there is ``love" in the generated sentence, the speaker still expresses his or her negative sentiment by irony. We also make some explorations in the transformation from ironic sentences to non-ironic sentences at the end of our work. Because of the lack of previous work and baselines on irony generation, we implement our model based on style transfer. Our work will not only provide the first large-scale irony dataset but also make our model as a benchmark for the irony generation.

Recently, unsupervised style transfer becomes a very popular topic. Many state-of-the-art studies try to solve the task with sequence-to-sequence (seq2seq) framework. There are three main ways to build up models. The first is to learn a latent style-independent content representation and generate sentences with the content representation and another style \cite{DBLP:conf/nips/ShenLBJ17,DBLP:conf/acl/TsvetkovBSP18}. The second is to directly transfer sentences from one style to another under the control of classifiers and reinforcement learning \cite{DBLP:conf/ijcai/LuoLZYCSS19}. The third is to remove style attribute words from the input sentence and combine the remaining content with new style attribute words \cite{DBLP:conf/naacl/LiJHL18,DBLP:conf/acl/LiWZXRSZ18}. The first method usually obtains better performances via adversarial training with discriminators. The style-independent content representation, nevertheless, is not easily obtained \cite{DBLP:conf/iclr/LampleSSDRB19}, which results in poor performances. The second method is suitable for complex styles which are difficult to model and describe. The model can learn the deep semantic features by itself but sometimes the model is sensitive to parameters and hard to train. The third method succeeds to preserve content but cannot work for some complex styles such as democratic and republican. Sentences with those styles usually do not have specific style attribute words. Unfortunately, due to the lack of large irony dataset and difficulties of modeling ironies, there has been little work trying to generate ironies based on seq2seq framework as far as we know. Inspired by methods for style transfer, we decide to implement a specifically designed model based on unsupervised style transfer to explore irony generation.

In this paper, in order to address the lack of irony data, we first crawl over 2M tweets from twitter to build a dataset with 262,755 ironic and 112,330 non-ironic tweets. Then, due to the lack of parallel data, we propose a novel model to transfer non-ironic sentences to ironic sentences in an unsupervised way. As ironic style is hard to model and describe, we implement our model with the control of classifiers and reinforcement learning. Different from other studies in style transfer, the transformation from non-ironic to ironic sentences has to preserve sentiment polarity as mentioned above. Therefore, we not only design an irony reward to control the irony accuracy and implement denoising auto-encoder and back-translation to control content preservation but also design a sentiment reward to control sentiment preservation.

Experimental results demonstrate that our model achieves a high irony accuracy with well-preserved sentiment and content. The contributions of our work are as follows:
\begin{itemize}
    \item To our knowledge, our work is the first attempt to specifically define irony generation and generate ironic sentences via seq2seq neural network.
    \item We build the first large-scale irony dataset with 262,755 ironic and 112,330 non-ironic tweets.
    \item We implement well-designed rewards for irony accuracy and sentiment preservation which are different from those in previous work on style transfer
    \item Our approach yields substantial results on generating ironic sentences with high irony accuracy and well-preserved content and sentiment.
\end{itemize}

\section{Related Work}
\textbf{Style Transfer}: 
As irony is a complicated style and hard to model with some specific style attribute words, we mainly focus on studies without editing style attribute words. 

Some studies are trying to disentangle style representation from content representation. 
In \cite{DBLP:conf/aaai/FuTPZY18}, authors leverage adversarial networks to learn separate content representations and style representations. In \cite{DBLP:conf/icml/HuYLSX17} and \cite{DBLP:conf/nips/ShenLBJ17}, researchers combine variational auto-encoders (VAEs) with style discriminators.

However,  some recent studies \cite{DBLP:conf/iclr/LampleSSDRB19} reveal that the disentanglement of content and style representations may not
be achieved in practice. Therefore, some other research studies \cite{DBLP:conf/naacl/LiJHL18,DBLP:conf/acl/LiWZXRSZ18} strive to separate content and style by removing stylistic words. Nonetheless, many non-ironic sentences do not have specific stylistic words and as a result, we find it difficult to transfer non-ironic sentences to ironic sentences through this way in practice.

Besides, some other research studies do not disentangle style from content but directly learn representations of sentences. In \cite{DBLP:conf/ijcai/LuoLZYCSS19}, authors propose a dual reinforcement learning framework without separating content and style representations. In \cite{DBLP:conf/acl/TsvetkovBSP18}, researchers utilize a machine translation model to learn a sentence representation preserving the meaning of the sentence but reducing stylistic properties. In this method, the quality of generated sentences relies on the performance of classifiers to a large extent. Meanwhile, such models are usually sensitive to parameters and difficult to train. In contrast, we combine a pre-training process with reinforcement learning to build up a stable language model and design special rewards for our task.

\noindent\textbf{Irony Detection}: With the development of social media, irony detection becomes a more important task. Methods for irony detection can be mainly divided into two categories: methods based on feature engineering and methods based on neural networks.

As for methods based on feature engineering, 
In \cite{DBLP:conf/eacl/MoriceauKBPBA17}, authors investigate pragmatic phenomena and various irony markers. In \cite{DBLP:conf/semeval/RohanianTEM18}, researchers leverage a combination of sentiment, distributional semantic and text surface features. Those models rely on hand-crafted features and are hard to implement.

When it comes to methods based on neural networks, long short-term memory (LSTM) \cite{DBLP:journals/neco/HochreiterS97} network is widely used and is very efficient for irony detection. In \cite{DBLP:conf/semeval/GhoshV18}, a tweet is divided into two segments and a subtract layer is implemented to calculate the difference between two segments in order to determine whether the tweet is ironic. In \cite{DBLP:conf/semeval/BaziotisNCKPENP18}, authors utilize a recurrent neural network with Bi-LSTM and self-attention without hand-crafted features. In \cite{DBLP:conf/semeval/WuWWLYH18}, researchers propose a system based on a densely connected LSTM network.

\begin{table*}[t]
\begin{center}
\caption{Samples in our dataset.}
\begin{tabular}{l | l} 
 \hline
  & Examples \\ [0.5ex] 
 \hline
  \multirow{2}{*}{Irony} & i love this time of year , when all your teacher throw in one more big project before finals .\\
  \cline{2-2}
     & i just love it when my plans get ruined .\\
 \hline
 \multirow{2}{*}{Non-irony} &  could be my last day in collingwood place . i am so sad .\\
 \cline{2-2}
           &  i hate my roommates for making drink on a tuesday .\\
 \hline
\end{tabular}    
\end{center}

\label{table:exp_dataset}
\end{table*}

\section{Our Dataset}
In this section, we describe how we build our dataset with tweets. First, we crawl over 2M tweets from twitter\footnote{https://twitter.com/.} using GetOldTweets-python\footnote{https://github.com/Jefferson-Henrique/GetOldTweets-python.}. We crawl English tweets from 04/09/2012 to /12/18/2018. We first remove all re-tweets and use langdetect\footnote{https://github.com/Mimino666/langdetect.} to remove all non-English sentences. Then, we remove hashtags attached at the end of the tweets because they are usually not parts of sentences and will confuse our language model. After that, we utilize Ekphrasis\footnote{https://github.com/cbaziotis/ekphrasis.} to process tweets. We remove URLs and restore remaining hashtags, elongated words, repeated words, and all-capitalized words. To simplify our dataset, We replace all ``$<$money$>$" and ``$<$time$>$" tokens with ``$<$number$>$" token when using Ekphrasis. And we delete sentences whose lengths are less than 10 or greater than 40. In order to restore abbreviations, we download an abbreviation dictionary from webopedia\footnote{https://www.webopedia.com/quick\_ref/ textmessageabbreviations.asp.} and restore abbreviations to normal words or phrases according to the dictionary. Finally, we remove sentences which have more than two rare words (appearing less than three times) in order to constrain the size of vocabulary. Finally, we get 662,530 sentences after pre-processing.

As neural networks are proved effective in irony detection, we decide to implement a neural classifier in order to classify the sentences into ironic and non-ironic sentences. However, the only high-quality irony dataset we can obtain is the dataset of Semeval-2018 Task 3 and the dataset is pretty small, which will cause overfitting to complex models. Therefore, we just implement a simple one-layer RNN with LSTM cell to classify pre-processed sentences into ironic sentences and non-ironic sentences because LSTM networks are widely used in irony detection. We train the model with the dataset of Semeval-2018 Task 3. After classification, we get 262,755 ironic sentences and 399,775 non-ironic sentences. According to our observation, not all non-ironic sentences are suitable to be transferred into ironic sentences. For example, ``just hanging out . watching . is it monday yet" is hard to transfer because it does not have an explicit sentiment polarity. So we remove all interrogative sentences from the non-ironic sentences and only obtain the sentences which have words expressing strong sentiments. We evaluate the sentiment polarity of each word with TextBlob\footnote{https://github.com/sloria/TextBlob.} and we view those words with sentiment scores greater than 0.5 or less than -0.5 as words expressing strong sentiments. Finally, we build our irony dataset with 262,755 ironic sentences and 102,330 non-ironic sentences.

\begin{figure}[t]
    \centering
    \includegraphics[scale=0.6]{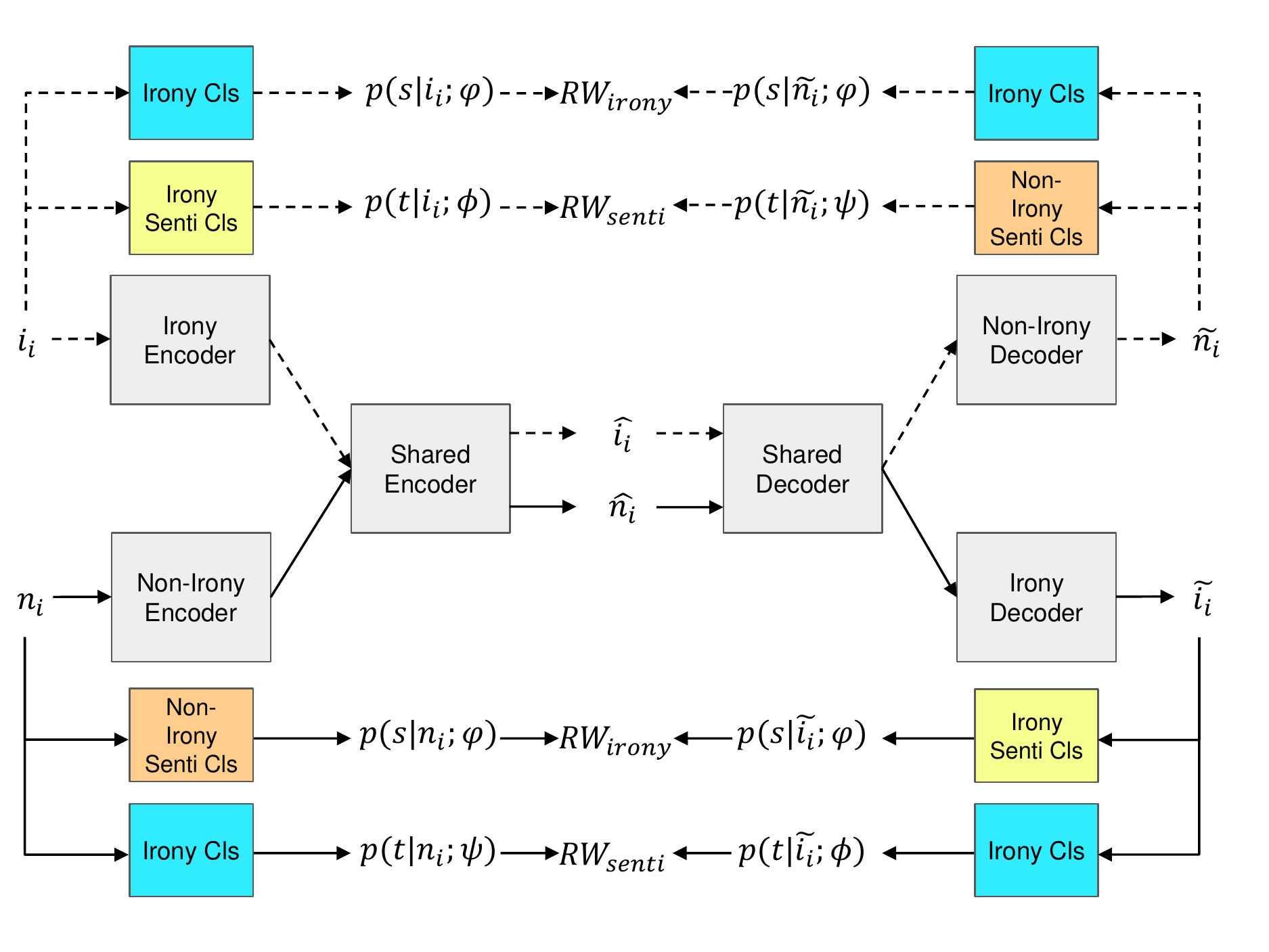}
    \caption{Framework of our model. The solid lines denote the transformation from non-ironic sentences to ironic sentences and the dotted lines denote the transformation from ironic sentences to non-ironic sentences.}
    \label{fig:1}
\end{figure}

\begin{algorithm}[t]
\caption{Irony Generation Algorithm}
\label{alg:overall}
\begin{algorithmic}
\footnotesize

\STATE $\triangleright$ pre-train with auto-encoder
\STATE Pre-train $E_n$, $D_n$ with $\textbf{\textit{N}}$ using MLE based on Eq.\ref{loss_ae_n}
\STATE Pre-train $E_i$, $D_i$ with $\textbf{\textit{I}}$ using MLE based on Eq.\ref{loss_ae_i}
\STATE $\triangleright$ pre-train with back-translation
\STATE Pre-train $E_n$, $D_n$, $E_i$, $D_i$ with $\textbf{\textit{N}}$ using MLE based on Eq.\ref{loss_bt_n}
\STATE Pre-train $E_n$, $D_n$, $E_i$, $D_i$ with $\textbf{\textit{I}}$ using MLE based on Eq.\ref{loss_bt_i}

\STATE
\STATE $\triangleright$ train with RL
\FOR{each epoch e = 1, 2, ..., $M$}
\STATE $\triangleright$ train non-irony2irony with RL
\FOR{$n_i$ in \textbf{\textit{N}}}
\STATE $\widetilde{i_i}=D_i(E_n(n_i))$
\STATE update $E_n$, $D_i$, using $Loss_{RL}$ based on Eq.\ref{loss_rl}
\STATE $\triangleright$ back-translation
\IF{$i\% p=0$}
\STATE $\widetilde{i_i}=D_i(E_n(n_i))$
\STATE $\widetilde{n_i}=D_n(E_i(\widetilde{i_i}))$
\STATE update $E_n$, $D_n$, $E_i$, $D_i$ using MLE based on Eq.\ref{loss_bt_n}
\ENDIF
\ENDFOR
\STATE $\triangleright$ train irony2non-irony with RL
\FOR{$i_i$ in \textbf{\textit{I}}}
\STATE $\widetilde{n_i}=D_n(E_i(i_i))$
\STATE update $E_i$, $D_n$, using $Loss_{RL}$ similar to Eq.\ref{loss_rl}
\STATE $\triangleright$ back-translation
\IF{$i\% p=0$}
\STATE $\widetilde{n_i}=D_n(E_i(i_i))$
\STATE $\widetilde{i_i}=D_i(E_n(\widetilde{n_i}))$
\STATE update $E_n$, $D_n$, $E_i$, $D_i$ using MLE based on Eq.\ref{loss_bt_i}
\ENDIF
\ENDFOR
\ENDFOR

\end{algorithmic}
\end{algorithm}

\section{Our Method}
Given two non-parallel corpora: non-ironic corpus \textbf{\textit{N}}=\{$n_1$, $n_2$, ..., $n_n$\} and ironic corpus \textbf{\textit{I}}=\{$i_1$, $i_2$, ..., $i_m$\}, the goal of our irony generation model is to generate an ironic sentence from a non-ironic sentence while preserving the content and sentiment polarity of the source input sentence. We implement an encoder-decoder framework where two encoders are utilized to encode ironic sentences and non-ironic sentences respectively and two decoders are utilized to decode ironic sentences and non-ironic sentences from latent representations respectively. In order to enforce a shared latent space, we share two layers on both the encoder side and the decoder side. Our model architecture is illustrated in Figure \ref{fig:1}. We denote irony encoder as $E_i$, irony decoder as $D_i$ and non-irony encoder as $E_n$, non-irony decoder as $D_n$. Their parameters are $\theta_{E_i}$, $\theta_{D_i}$, $\theta_{E_n}$ and $\theta_{D_n}$.

Our irony generation algorithm is shown in Algorithm \ref{alg:overall}. We first pre-train our model using denoising auto-encoder and back-translation to build up language models for both styles (section \ref{subsec: pretraining}). Then we implement reinforcement learning to train the model to transfer sentences from one style to another (section \ref{subsec: rl}). Meanwhile, to achieve content preservation, we utilize back-translation for one time in every $p$ time steps.

\subsection{Pretraining}
\label{subsec: pretraining}
\subsubsection{Denoising Auto-Encoder}
In order to build up our language model and preserve the content, we apply the  auto-encoder model. To prevent the model from simply copying the input sentence, we randomly add some noises in the input sentence. Specifically, for every word in the input sentence, there is 10\% chance that we delete it, 10 \% chance that we duplicate it, 10\% chance that we swap it with the next word, or it remains unchanged. We first encode the input sentence $n_i$ or $i_i$ with respective encoder $E_n$ or $E_i$ to obtain its latent representation $\hat{n_i}=E_n(n_i)$ or $\hat{i_i}=E_i(i_i)$ and reconstruct the input sentence with the latent representation and respective decoder. So we can get the reconstruction loss for auto-encoder $Loss_{ae}$:
\begin{equation}
Loss_{ae}^n=\mathbb{E}_{n\sim\textbf{\textit{N}}}[-\log p_{D_n}(n|\hat{n_i};\theta_{D_n})]
\label{loss_ae_n}
\end{equation}
\begin{equation}
Loss_{ae}^i=\mathbb{E}_{i\sim\textbf{\textit{I}}}[-\log p_{D_i}(i|\hat{i_i};\theta_{D_i})]
\label{loss_ae_i}
\end{equation}

\subsubsection{Back Translation}
In addition to denoising auto-encoder, we implement back-translation \cite{DBLP:conf/acl/SennrichHB16} to generate a pseudo-parallel corpus. Suppose our model takes non-ironic sentence $n_i$ as input. We first encode $n_i$ with $E_n$ to obtain its latent representation $\hat{n_i}=E_n(n_i)$ and decode the latent representation with $D_i$ to get a transferred sentence $\widetilde{i_i}$. Then we encode $\widetilde{i_i}$ with $E_i$ and decode its latent representation with $D_n$ to reconstruct the original input sentence $n_i$. Therefore, our reconstruction loss for back-translation $Loss_{bt}$:
\begin{equation}
    Loss_{bt}^n=\mathbb{E}_{n\sim\textbf{\textit{N}}}[-\log p_{D_n}(n|E_i(\widetilde{i_i});\theta_{D_n})]
\label{loss_bt_n}
\end{equation}
And if our model takes ironic sentence $i_i$ as input, we can get the reconstruction loss for back-translation as:
\begin{equation}
    Loss_{bt}^i=\mathbb{E}_{i\sim\textbf{\textit{I}}}[-\log p_{D_i}(i|E_n(\widetilde{n_i});\theta_{D_i})]
\label{loss_bt_i}
\end{equation}

\subsection{Reinforcement Learning}
\label{subsec: rl}
Since the gold transferred result of input is unavailable, we cannot evaluate the quality of the generated sentence directly. Therefore, we implement reinforcement learning and elaborately design two rewards to describe the irony accuracy and sentiment preservation, respectively.
\subsubsection{Irony Reward}
A pre-trained binary irony classifier based on CNN \cite{DBLP:conf/emnlp/Kim14} is used to evaluate how ironic a sentence is. We denote the parameter of the classifier as $\varphi$ and it is fixed during the training process.

In order to facilitate the transformation, we design the irony reward as the difference between the irony score of the input sentence and that of the output sentence. Formally, when we input a non-ironic sentence $n_i$ and transfer it to an ironic sentence $\widetilde{i_i}$, our irony reward is defined as:
\begin{equation}
    RW_{irony}=p(s|\widetilde{i_i};\varphi)-p(s|n_i;\varphi)
\end{equation}
where $s$ denotes ironic style and $p(s|x;\varphi)$ is the probability of that a sentence $x$ is ironic.

\subsubsection{Sentiment Reward}
To preserve the sentiment polarity of the input sentence, we also need to use classifiers to evaluate the sentiment polarity of the sentences. However, the sentiment analysis of ironic sentences and non-ironic sentences are different. In the case of figurative languages such as irony, sarcasm or metaphor, the sentiment polarity of the literal meaning may differ significantly from that of the intended figurative
meaning \cite{DBLP:conf/semeval/GhoshLVRSBR15}. As we aim to train our model to transfer sentences from non-ironic to ironic, using only one classifier is not enough. As a result, we implement two pre-trained sentiment classifiers for non-ironic sentences and ironic sentences respectively. We denote the parameter of the sentiment classifier for ironic sentences as $\phi$ and that of the sentiment classifier for non-ironic sentences as $\psi$. 

A challenge, when we implement two classifiers to evaluate the sentiment polarity, is that the two classifiers trained with different datasets may have different distributions of scores. That means we cannot directly calculate the sentiment reward with scores applied by two classifiers. To alleviate this problem and standardize the prediction results of two classifiers, we set a threshold for each classifier and subtract the respective threshold from scores applied by the classifier to obtain the comparative sentiment polarity score. We get the optimal threshold by maximizing the ability of the classifier according to the distribution of our training data.

We denote the threshold of ironic sentiment classifier as $th_i$ and the threshold of non-ironic sentiment classifier as $th_n$. The standardized sentiment score is defined as $STD(p(t|n_i;\psi))=p(t|n_i;\psi)-th_n$ and $STD(p(t|i_i;\phi))=p(t|i_i;\phi)-th_i$ where $t$ denotes the positive sentiment polarity and $p(t|x;\cdot)$ is the probability of that a sentence is positive in sentiment polarity.

As mentioned above, the input sentence and the generated sentence should express the same sentiment. For example, if we input a non-ironic sentence ``I hate to be ignored" which is negative in sentiment polarity, the generated ironic sentence should be also negative, such as ``I love to be ignored". To achieve sentiment preservation, we design the sentiment reward as that one minus the absolute value of the difference between the standardized sentiment score of the input sentence and that of the generated sentence. Formally, when we input a non-ironic sentence $n_i$ and transfer it to an ironic sentence $\widetilde{i_i}$, our sentiment reward is defined as:
\begin{equation}
    RW_{senti}=1-abs(STD(p(t|n_i;\psi))-STD(p(t|\widetilde{i_i};\phi)))
\label{eq:senti_rw_n2i}
\end{equation}

\subsubsection{Overall Reward}
To encourage our model to focus on both the irony accuracy and the sentiment preservation, we apply the harmonic mean of irony reward and sentiment reward:
\begin{equation}
    RW = (1+\beta^2)\frac{RW_{senti}\cdot RW_{irony}}{(\beta^2\cdot RW_{senti}) + RW_{irony}}
\label{rw_overall}
\end{equation}

\subsection{Policy Gradient}
The policy gradient algorithm \cite{DBLP:journals/ml/Williams92} is a simple but widely-used algorithm in reinforcement learning. It is used to maximize the expected reward $\mathbb{E}[RW]$. The objective function to minimize is defined as:
\begin{equation}
    Loss_{RL}=-\frac{1}{K}\sum_{i=1}^{K}RW_i\cdot p(\widetilde{i_i}|n_i;\theta_{E_n},\theta_{D_i})
\label{loss_rl}
\end{equation}
where $\widetilde{i_i}=D_i(E_n(n_i))$, $RW_i$ is the reward of $\widetilde{i_i}$ and $K$ is the input size.

\section{Experiments}
\subsection{Training Details}
$E_x$, $E_y$, $D_x$ and $D_y$ in our model are Transformers \cite{DBLP:conf/nips/VaswaniSPUJGKP17} with 4 layers and 2 shared layers. The word embeddings of 128 dimensions are learned during the training process. Our maximum sentence length is set as 40. The optimizer is Adam \cite{DBLP:journals/corr/KingmaB14} and the learning rate is $10^{-5}$. The batch size is 32 and harmonic weight $\beta$ in Eq.9 is 0.5. We set the interval $p$ as 200. The model is pre-trained for 6 epochs and trained for 15 epochs for reinforcement learning. 

\begin{itemize}
    \item Irony Classifier: We implement a CNN classifier trained with our irony dataset. All the CNN classifiers we utilize in this paper use the same parameters as \cite{DBLP:conf/emnlp/Kim14}.
    \item Sentiment Classifier for Irony: We first implement a one-layer LSTM network to classify ironic sentences in our dataset into positive and negative ironies. The LSTM network is trained with the dataset of Semeval 2015 Task 11 \cite{DBLP:conf/semeval/GhoshLVRSBR15} which is used for the sentiment analysis of figurative language in twitter. Then, we use the positive ironies and negative ironies to train the CNN sentiment classifier for irony.
    \item Sentiment Classifier for Non-irony: Similar to the training process of the sentiment classifier for irony, we first implement a one-layer LSTM network trained with the dataset for the sentiment analysis of common twitters\footnote{https://www.kaggle.com/c/twitter-sentiment-analysis2/data.} to classify the non-ironies into positive and negative non-ironies. Then we use the positive and negative non-ironies to train the sentiment classifier for non-irony.
\end{itemize}

\begin{table}[t]
\begin{center}

\caption{Automatic evaluation results of the transformation from non-ironic sentences to ironic sentences.\textsuperscript{*}}
\label{table:auto_n2i}
\begin{tabular}{lrrrrr}
     \toprule
     Model & Senti Delta$\downarrow$ & Senti ACC$\uparrow$ &  BLEU$\uparrow$ & \textbf{G2}$\uparrow$ & \textbf{H2}$\uparrow$ \\
     \midrule 
     BackTrans & 0.5417 & 48.83 & 1.80 & 9.38 & 3.47\\
     Unpaired & 0.5454 & 49.26 & 18.78 & 30.41 & 27.19\\
     CrossAlign & 0.5357 & 49.56 & 2.77 & 11.72 & 5.25\\
     CPTG & 0.5174 & 49.43 & 0.26 & 3.58 & 0.52\\
     DualRL & 0.5167 & \textbf{49.73} & \textbf{76.38} & \textbf{61.63} & \textbf{60.24}\\
     \midrule 
     Ours & \textbf{0.5148} & 49.68 & 61.78 & 55.40 & 55.07\\
     \bottomrule
\end{tabular}
\end{center}

\small\textsuperscript{*} We use $\downarrow$ to denote that the smaller value is better and $\uparrow$ to denote that the larger value is better.

\end{table}

\begin{table}[t]
\begin{center}
\caption{Human evaluation results of the transformation from non-ironic sentences to ironic sentences.}
\label{table:human_n2i}
\begin{tabular}{lrrr}
     \toprule
     Model & Irony$\downarrow$ &  Senti$\downarrow$  & Content$\downarrow$  \\
     \midrule 
     BackTrans & 2.98 & 3.86 & 3.91\\
     Unpaired & 4.35 & 4.26 & 4.05\\
     CrossAlign & 3.72 & 4.26 & 4.30\\
     CPTG & 4.65 & 4.84 & 5.00\\
     DualRL & 2.91 & \textbf{1.28} & \textbf{1.35}\\
     \midrule 
     Ours & \textbf{2.40} & 2.51 & 2.40\\
     \bottomrule
\end{tabular}
\end{center}

\end{table}

\begin{table*}[t]
\begin{center}
\caption{Example outputs of the transformation from non-ironic sentences to ironic sentences and the transformation from ironic sentences to non-ironic sentences. We use \textcolor{red}{red} and \textcolor{blue}{blue} to annotate the clashes in the sentences.}
\label{table:exp_outputs_n2i}
\scalebox{0.66}{
\begin{tabular}{l|l|l}
     \toprule
      & Example 1 (From non-irony to irony) & Example 2 (From non-irony to irony)  \\
     \midrule 
     source & tried to leave town and my phone died , definition of success . & not even sleepy . tomorrows going to be fun .\\
     \midrule
     BackTrans & getting up at <number> in the morning and my life is going to be so much fun . &  not being so much fun . i am gonna be a great day .\\
     Unpaired & tried to leave town and my phone died , plowing split xx . &  no gets cool .\\
     CrossAlign & going to work and my house and <number> minutes , &   can not wait to go . this is going to work .\\
     CPTG & very little <UNK> on <user> , \& you like cold tower getting good not ppl know . & not people but $<$user$>$ if you are binge - of way tell he " on festivities right .\\
     DualRL & tried to leave town and my phone died , definition of success . & not even sleepy . tomorrows going to be fun .\\
     \midrule 
     Ours & \textcolor{blue}{nice} to leave town and my phone \textcolor{red}{died} , definition of success . & not even sleepy . \textcolor{red}{depressed} going to be \textcolor{blue}{fun} .\\
     \bottomrule
\end{tabular}
}
\end{center}

\end{table*}

\begin{table*}[t]
\begin{center}
\caption{Example error outputs of the transformation from non-ironic sentences to ironic sentences. The main errors are \textcolor{red}{colored}.}
\label{table:err_outputs}
\scalebox{0.72}{
\begin{tabular}{l|l|l}
     \toprule
      & source (non-irony) & output (irony) \\
     \midrule
     No Change & cool thanks for even coming to say hi to me . & cool thanks for even coming to say hi to me .\\
     \midrule 
     Word Repetition & not able to extract data from tabs using python path
 & not able to \textcolor{red}{accountable accountable} from \textcolor{red}{accountable} using \textcolor{red}{accountable} accountable\\
    \midrule
    Improper Words & friday night and here i am playing fifa . wonderful & \textcolor{red}{feel} night and here i am playing \textcolor{red}{just} . wonderful\\
     \bottomrule
\end{tabular}}
\end{center}

\end{table*}

\subsection{Baselines}
We compare our model with the following state-of-art generative models:\\
\noindent\textbf{BackTrans\cite{DBLP:conf/acl/TsvetkovBSP18}:} In \cite{DBLP:conf/acl/TsvetkovBSP18}, authors propose a model using machine translation in order to preserve the meaning of the sentence while reducing stylistic properties.\\
\noindent\textbf{Unpaired\cite{DBLP:conf/acl/LiWZXRSZ18}: } In \cite{DBLP:conf/acl/LiWZXRSZ18}, researchers implement a method to remove emotional words and add desired sentiment controlled by reinforcement learning.\\
\noindent\textbf{CrossAlign\cite{DBLP:conf/nips/ShenLBJ17}:}  In \cite{DBLP:conf/nips/ShenLBJ17}, authors leverage refined alignment of latent representations to perform style transfer and a cross-aligned auto-encoder is implemented.\\
\noindent\textbf{CPTG\cite{DBLP:conf/nips/LogeswaranLB18}:} An interpolated reconstruction loss is introduced in \cite{DBLP:conf/nips/LogeswaranLB18} and a discriminator is implemented to control attributes in this work.\\
\noindent\textbf{DualRL\cite{DBLP:conf/ijcai/LuoLZYCSS19}:} In \cite{DBLP:conf/ijcai/LuoLZYCSS19}, researchers use two reinforcement rewards simultaneously to control style accuracy and content preservation.

\subsection{Evaluation Metrics}
\subsubsection{Automatic Evaluation}
In order to evaluate sentiment preservation, we use the absolute value of the difference between the standardized sentiment score of the input sentence and that of the generated sentence\footnote{See the definition in Eq. \ref{eq:senti_rw_n2i}. As sentiment reward, we use one to minus the absolute value and now we only use the absolute value as the metric.}. We call the value as sentiment delta (senti delta). Besides, we report the sentiment accuracy (Senti ACC) which measures whether the output sentence has the same sentiment polarity as the input sentence based on our standardized sentiment classifiers. The BLEU score \cite{DBLP:conf/acl/PapineniRWZ02} between the input sentences and the output sentences is calculated to evaluate the content preservation performance. In order to evaluate the overall performance of different models, we also report the geometric mean (G2) and harmonic mean (H2) of the sentiment accuracy and the BLEU score. As for the irony accuracy, we only report it in human evaluation results because it is more accurate for the human to evaluate the quality of irony as it is very complicated.

\subsubsection{Human Evaluation}
We first sample 50 non-ironic input sentences and their corresponding output sentences of different models. Then, we ask four annotators who are proficient in English to evaluate the qualities of the generated sentences of different models. They are required to rank the output sentences of our model and baselines from the best to the worst in terms of irony accuracy (Irony), Sentiment preservation (Senti) and content preservation (Content). The best output is ranked with 1 and the worst output is ranked with 6. That means that the smaller our human evaluation value is, the better the corresponding model is. 

\subsection{Results and Discussions}
Table \ref{table:auto_n2i} shows the automatic evaluation results of the models in the transformation from non-ironic sentences to ironic sentences. From the results, our model obtains the best result in sentiment delta. The DualRL model achieves the highest result in other metrics, but most of its outputs are the almost same as the input sentences. So it is reasonable that DualRL system outperforms ours in these metrics but it actually does not transfer the non-ironic sentences to ironic sentences at all. From this perspective, we cannot view DualRL as an effective model for irony generation. In contrast, our model gets results close to those of DualRL and obtains a balance between irony accuracy, sentiment preservation, and content preservation if we also consider the irony accuracy discussed below.

And from human evaluation results shown in Table \ref{table:human_n2i}, our model gets the best average rank in irony accuracy. And as mentioned above, the DualRL model usually does not change the input sentence and outputs the same sentence. Therefore, it is reasonable that it obtains the best rank in sentiment and content preservation and ours is the second. However, it still demonstrates that our model, instead of changing nothing, transfers the style of the input sentence with content and sentiment preservation at the same time.

\subsection{Case Study}
In the section, we present some example outputs of different models. Table \ref{table:exp_outputs_n2i} shows the results of the transformation from non-ironic sentences to ironic sentences. We can observe that: (1) The BackTrans system, the Unpaired system, the CrossAlign system and the CPTG system tends to generate sentences which are towards irony but do not preserve content. (2) The DualRL system preserves content and sentiment very well but even does not change the input sentence. (3) Our model considers both aspects and achieves a better balance among irony accuracy, sentiment and content preservation.

\begin{table}[t]
\begin{center}
\caption{Automatic evaluation results of the transformation from ironic sentences to non-ironic sentences.}
\label{table:auto_i2n}
\begin{tabular}{lrrrrr}
     \toprule
     Model & Senti Delta$\downarrow$ & Senti ACC$\uparrow$ & BLEU$\uparrow$ & \textbf{G2}$\uparrow$ & \textbf{H2}$\uparrow$ \\
     \midrule 
     BackTrans & 0.6927 & 40.87 & 1.98 & 9.00 & 3.78\\
     Unpaired & 0.5169 & \textbf{49.64} & 9.28 & 21.46 & 15.64\\
     CrossAlign & 0.5710 & 46.77 & 4.85 & 15.06 & 8.79\\
     CPTG & 0.5172 & 48.94 & 0.49 & 4.90 & 0.97\\
     DualRL & 0.5534 & 47.82 & \textbf{74.31} & \textbf{59.61} & \textbf{58.19}\\
     \midrule 
     Ours & \textbf{0.4976} & 49.09 & 62.92 & 57.33 & 56.64\\
     \bottomrule
\end{tabular}
\end{center}

\end{table}

\begin{table}[t]
\begin{center}
\caption{Human evaluation results of the transformation from ironic sentences to non-ironic sentences. Note that the larger value for irony metric is better here as we generate non-ironic sentences.}
\label{table:human_i2n}
\begin{tabular}{lrrr}
     \toprule
     Model & Irony$\uparrow$ &  Senti$\downarrow$  & Content$\downarrow$  \\
     \midrule 
     BackTrans & 3.46 & 3.73 & 3.94\\
     Unpaired & 4.00 & 3.94 & 3.82\\
     CrossAlign & 3.68 & 3.99 & 4.27\\
     CPTG & \textbf{4.86} & 5.07 & 5.00\\
     DualRL & 1.93 & \textbf{1.60} & \textbf{1.43}\\
     \midrule 
     Ours & 3.07 & 2.67 & 2.54\\
     \bottomrule
\end{tabular}
\end{center}

\end{table}

\subsection{Error Analysis}
Although our model outperforms other style transfer baselines according to automatic and human evaluation results, there are still some failure cases because irony generation is still a very challenging task. We would like to share the issues we meet during our experiments and our solutions to some of them in this section.

\begin{itemize}
    \item No Change: As mentioned above, many style transfer models, such as DualRL, tend to make few changes to the input sentence and output the same sentence. Actually, this is a common issue for unsupervised style transfer systems and we also meet it during our experiments. The main reason for the issue is that rewards for content preservation are too prominent and rewards for style accuracy cannot work well. In contrast, in order to guarantee the readability and fluency of the output sentence, we also cannot emphasize too much on rewards for style accuracy because it may cause some other issues such as word repetition mentioned below. A method to solve the problem is tuning hyperparameters and this is also the method we implement in this work. As for content preservation, maybe MLE methods such as back-translation are not enough because they tend to force models to generate specific words. In the future, we should further design some more suitable methods to control content preservation for models without disentangling style and content representations, such as DualRL and ours. 
    \item Word Repetition: During our experiments, we observe that some of the outputs prefer to repeat the same word as shown in Table \ref{table:err_outputs}. This is because reinforcement learning rewards encourage the model to generate words which can get high scores from classifiers and even back-translation cannot stop it. Our solution is that we can lower the probability of decoding a word in decoders if the word has been generated in the previous time steps during testing. We also try to implement this method during training time but obtain worse performances because it may limit the effects of training. Some previous studies utilize language models to control the fluency of the output sentence and we also try this method. Nonetheless, pre-training a language model with tweets and using it to generate rewards is difficult because tweets are more casual and have more noise. Rewards from that kind of language model are usually not accurate and may confuse the model. In the future, we should come up with better methods to model language fluency with the consideration of irony accuracy, sentiment and content preservation, especially for tweets. 
    \item Improper Words: As ironic style is hard for our model to learn, it may generate some improper words which make the sentence strange. As the example shown in the Table \ref{table:err_outputs}, the sentiment word in the input sentence is ``wonderful" and the model should change it into a negative word such as ``sad" to make the output sentence ironic. However, the model changes ``friday" and ``fifa" which are not related to ironic styles. We have not found a very effective method to address this issue and maybe we should further explore stronger models to learn ironic styles better. 
\end{itemize}

\begin{table}[t]
\begin{center}
\caption{Example outputs of the transformation from ironic sentences to non-ironic sentences. We use \textcolor{red}{red} and \textcolor{blue}{blue} to annotate the clashes in the sentences.}
\label{table:exp_outputs_i2n}
\scalebox{0.85}{
\begin{tabular}{l|l}
     \toprule
      & From ironic to non-ironic \\
     \midrule 
     source & wow i \textcolor{red}{love} feeling like the biggest \textcolor{blue}{dumbass} on earth .\\
     \midrule
     BackTrans &  $<$user$>$ i am sure you are going to be a good day .\\
     Unpaired &  i am good like always dude [UNK] on secret piece my\\
     CrossAlign &   i am so excited for the house of the house .\\
     CPTG & it is only i the hated senate meet house of this funny right now\\ &tonight outage\\
     DualRL & wow i \textcolor{red}{love} feeling like the biggest \textcolor{blue}{dumbass} on earth .\\
     \midrule 
     Ours & wow i started feeling like the biggest \textcolor{blue}{dumbass} on earth . \\
     \bottomrule
\end{tabular}}
\end{center}

\end{table}

\subsection{Additional Experiments}
In this section, we describe some additional experiments on the transformation from ironic sentences to non-ironic sentences. Sometimes ironies are hard to understand and may cause misunderstanding, for which our task also explores the transformation from ironic sentences to non-ironic sentences.

As shown in Table \ref{table:auto_i2n}, we also conduct automatic evaluations and the conclusions are similar to those of the transformation from non-ironic sentences to ironic sentences. 
As for human evaluation results in Table \ref{table:human_i2n}, our model still can achieve the second-best results in sentiment and content preservation. Nevertheless, DualRL system and ours get poor performances in irony accuracy. The reason may be that the other four baselines tend to generate common and even not fluent sentences which are irrelevant to the input sentences\footnote{See the examples in Table \ref{table:exp_outputs_n2i}.} and are hard to be identified as ironies. So annotators usually mark these output sentences as non-ironic sentences, which causes these models to obtain better performances than DualRL and ours but much poorer results in sentiment and content preservation. Some examples are shown in Table \ref{table:exp_outputs_i2n}.

\section{Conclusion and Future Work}
In this paper, we first systematically define irony generation based on style transfer. Because of the lack of irony data, we make use of twitter and build a large-scale dataset. In order to control irony accuracy, sentiment preservation and content preservation at the same time, we also design a combination of rewards for reinforcement learning and incorporate reinforcement learning with a pre-training process. Experimental results demonstrate that our model outperforms other generative models and our rewards are effective.
Although our model design is effective, there are still many errors and we systematically analyze them. In the future, we are interested in exploring these directions and our work may extend to other kinds of ironies which are more difficult to model. 

\bibliographystyle{unsrt}  
\bibliography{references}  

\end{document}